\newif\ifisTR

\isTRtrue

\documentclass{article} % For LaTeX2e
\usepackage{fullpage}

\usepackage{microtype}
\usepackage{graphicx}
\usepackage{booktabs} % for professional tables
\usepackage[dvipsnames]{xcolor}

\definecolor{mydarkblue}{rgb}{0,0.08,0.45}
\usepackage[colorlinks,citecolor=mydarkblue,urlcolor=mydarkblue,linkcolor=mydarkblue]{hyperref}

\usepackage{placeins}
\usepackage{algorithm}
\usepackage{algorithmic}
\usepackage{subcaption}
\usepackage{tcolorbox}
\usepackage{nicefrac}  
\usepackage{enumitem}
\usepackage{hyperref}
\usepackage{multirow}
\usepackage{colortbl}
\usepackage{xspace}
\usepackage{wrapfig}
\usepackage{amsmath,amsfonts,bm}

\def\eqref#1{equation~(\ref{#1})}
\def\1{\bm{1}}

\DeclareMathAlphabet{\mathsfit}{\encodingdefault}{\sfdefault}{m}{sl}
\SetMathAlphabet{\mathsfit}{bold}{\encodingdefault}{\sfdefault}{bx}{n}

\usepackage{amsmath}
\usepackage{amssymb}
\usepackage{mathtools}
\usepackage[export]{adjustbox}
\usepackage{amsthm}
\usepackage{xcolor}
\newcommand{\VP}[1]{{\color{black}#1}} %red
\usepackage{tikz}

\usepackage[capitalize,noabbrev]{cleveref}

\theoremstyle{plain}

\theoremstyle{definition}

\theoremstyle{remark}

\definecolor{mygray}{gray}{0.85}
\definecolor{LightBlue}{cmyk}{0.06, 0.03, 0.01, 0.0}

\usepackage[textsize=tiny]{todonotes}

\usepackage[round,semicolon,sort]{natbib}

\renewcommand{\cite}[1]{\citep{#1}}

\begin{document}

\title{When Does Visual Prompting Outperform Linear Probing for Vision-Language Models? A Likelihood Perspective}
\date{}

\author{%
 Hsi-Ai Tsao$^{1}$, 
  Lei Hsiung$^{2}$, 
  Pin-Yu Chen$^{3}$, 
  Tsung-Yi Ho$^{4}$ \\
  $^1$ National Tsing Hua University \\
  $^2$ Dartmouth College \\
  $^3$ IBM Research \\
  $^4$ The Chinese University of Hong Kong
}

\maketitle

\begin{abstract}
Adapting pre-trained models to new tasks can exhibit varying effectiveness across datasets. Visual prompting, a state-of-the-art parameter-efficient transfer learning method,
can significantly improve the performance of out-of-distribution tasks. On the other hand, linear probing, a standard transfer learning method, can sometimes become the best approach.
We propose a log-likelihood ratio (LLR) approach to analyze the comparative benefits of visual prompting and linear probing. By employing the LLR score alongside resource-efficient visual prompts approximations, our cost-effective measure attains up to a 100-fold reduction in run time compared to full training, while achieving prediction accuracies up to 91\%. The source code is available at \href{https://github.com/IBM/VP-LLR}{\texttt{VP-LLR}}.
\end{abstract}

\section{Introduction} \label{sec:intro}
When applying transfer learning to downstream tasks, specific modifications to the pre-trained model are required. For instance, linear probing (LP) involves adjusting the linear layer in the model’s penultimate layer, while full fine-tuning involves modifying all parameters in the model. However, in the emerging field of fine-tuning for transfer learning, visual prompting (VP) \cite{chen2024model, bahng2022exploring}  offers a method that does not necessitate changes to the pre-trained model.

Specifically, studies such as CLIP-VP \cite{bahng2022exploring} and AutoVP \cite{tsao2024autovp} indicate that visual prompting is particularly suitable for out-of-distribution (OOD) datasets. In AutoVP, the authors observed that datasets with lower confidence scores, indicative of being more OOD, tend to achieve greater \textbf{accuracy gains} (i.e., the performance difference between VP and LP).

\begin{figure}[tbp]
    \centering
    \includegraphics[width=\linewidth]{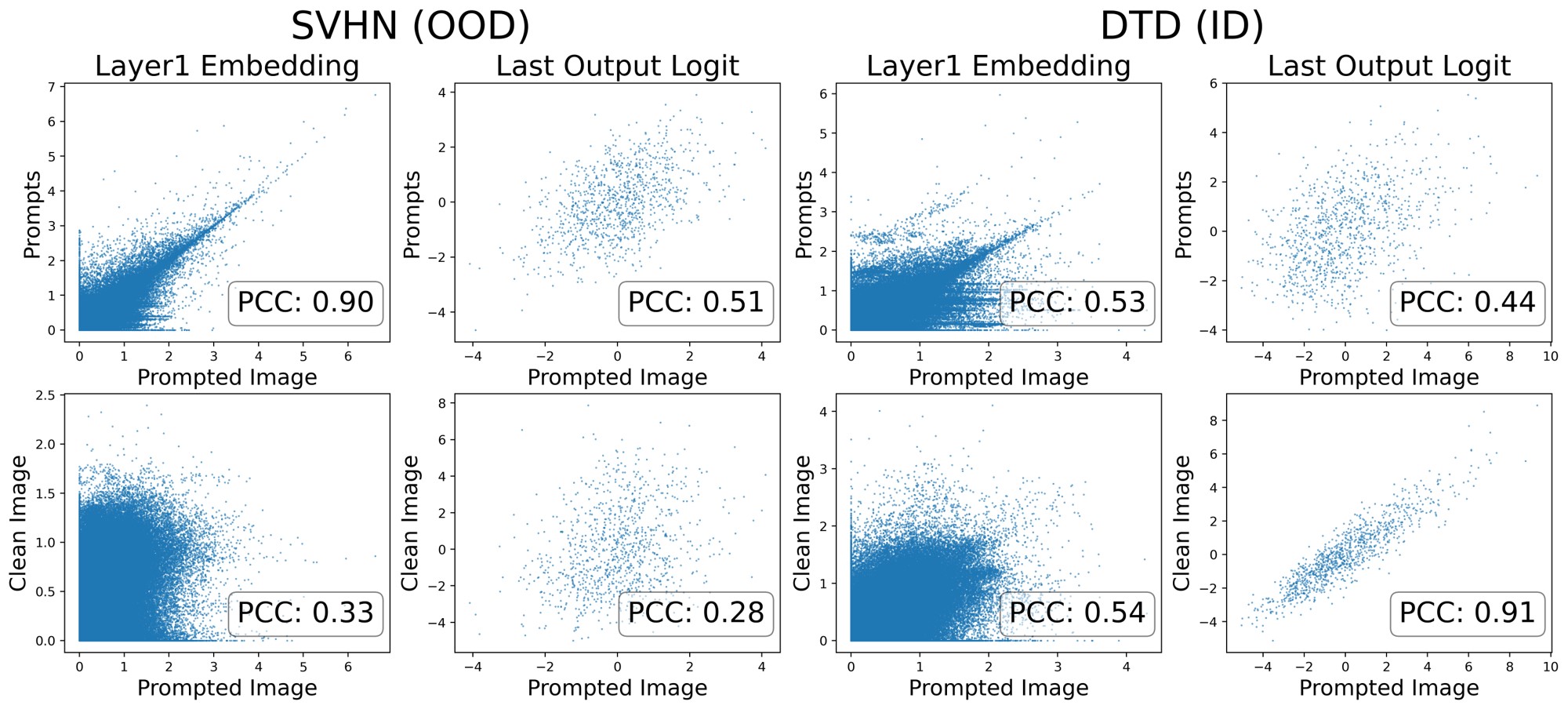}
    \caption {\textbf{The PCC of Embeddings in ResNet18.} The $PCC_{X,Y}$ is calculated by embeddings $X$ and $Y$. Here, $X$ is obtained by inputting the entire prompted image, while $Y$ is obtained by inputting either (1) the visual prompts or (2) the clean image.}
    \label{img:pcc}
\end{figure}

In this paper, we conduct an in-depth analysis of the effects of visual prompts on both OOD and in-distribution (ID) datasets. In Fig. \ref{img:pcc}, we compute the Pearson correlation coefficient (PCC) \cite{cohen2009pearson,zhang2020understanding} of model embeddings. The PCC is computed from embeddings generated by inputting the entire prompted image versus the image or prompt separately. The result (prompts or images) with a higher PCC score indicates greater similarity to the embedding obtained from the full input (i.e., the prompted image), indicating that it provides the dominant feature. The results show that in OOD datasets, prompts achieve a PCC of 0.9 in shallow layers, while in ID datasets, the output logits of clean images reach a PCC of 0.91. This implies that OOD datasets are better suited for training with VP, while ID datasets, to avoid interference with the inherent features of the image, are more appropriate for training with LP.

Since there is no one-fits-all method, selecting an applicable transfer learning method for downstream datasets remains critical. Some training-free approaches can serve as reliable references for estimating models' adaptability to downstream datasets, thereby preventing the need to explore the large space of the training configurations. Several studies have focused on pre-trained model selection \cite{nguyen2020leep, tran2019transferability, you2021logme}. Building on this idea, we extend LogME \cite{ you2021logme} to the selection of methods, VP or LP, providing a log-likelihood ratio (LLR) method as described in Section \ref{sec:llr}. Table \ref{tab:computation} presents the execution time across various methods, with an overall speedup of 100 times compared to linear probing (LP).

We summarize the main contributions as follows:
\begin{itemize}
\item We propose a cost-effective log-likelihood ratio (LLR) method to estimate whether VP or LP offers greater advantages on a given dataset.
\item The LLR scores effectively reflect the proportions of ID/OOD data in the dataset and align well with the accuracy differences between VP and LP.
\item The comprehensive results on 12 datasets (Section \ref{sec:llr_sorting}) using LLR scores outperform OOD detection baselines.
\end{itemize}

\section{Background and Related Work}\label{sec:related_work}
\subsection{Visual Prompting}
\begin{figure}[h]
    \centering
    \includegraphics[width=0.8\linewidth]{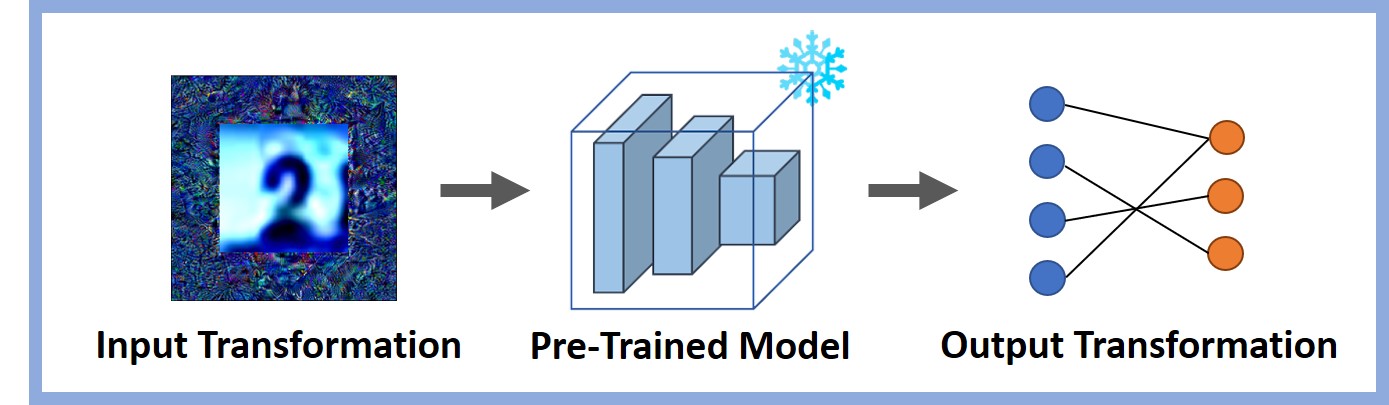} 
    \caption {\textbf{The Visual Prompting Framework.}}
    \label{img:vp_framework}
\end{figure}

Visual prompting (VP), also known as model reprogramming \cite{elsayed2018adversarial}, can be used to adapt a pre-trained model to new tasks. The VP framework, as shown in Fig. \ref{img:vp_framework}, consists of three components: input transformation, pre-trained models, and output transformation. In the input transformation, a trainable visual prompt is added, typically in the form of a frame padded around the image. The pre-trained model serves as a feature extractor and remains frozen during VP training. The output transformation then maps the pre-trained model's source labels to the target labels of the downstream task. Previous studies have investigated various VP designs and usage scenarios, including exploration of optimal prompt sizes \cite{bahng2022exploring, tsao2024autovp}, visual prompt tuning in vision transformers \cite{jia2022visual}, black-box VP training \cite{tsai2020transfer}, and iterative approaches to learning output mappings \cite{chen2023understanding}. These studies have demonstrated the capability and computational efficiency of VP.

\subsection{LogME for Model Selection} \label{related_work:logme}
Before performing model finetuning, maximum likelihood estimation methods can be used to select the most suitable pre-trained model for training. LogME \cite{you2021logme} employs evidence (i.e. marginalized likelihood) to assess the predictive capability of a model. The evidence is described as follows:
\begin{equation}
p(y|F,\alpha,\beta)=\int_{w} p(w|\alpha)p(y|F, w,\beta)dw
\end{equation}
where $y$ is the target label, $w$ follows an isotropic multivariate Gaussian distribution, representing the parameter of the linear layer added on top of the pre-trained model. $F = \{f_{i}\in \mathbb{R}^{D}\}_{i=1}^{n}$ is the feature matrix of dimension $D$ obtained from the model's last layer, and $\alpha^{-1}$ and $\beta^{-1}$ are the variances of $w$ and the prediction $y$, respectively. The log-likelihood can be further derived as follows \cite{you2021logme,bishop2006pattern}, where $A=\alpha I+\beta F^{T}F$ and $m=\beta A^{-1}F^{T}y$:
\begin{equation} \label{eq:LogME}
\begin{split}
\log(p(y|F,\alpha,\beta))&=\frac{D}{2}log(\alpha)+\frac{n}{2}log(\beta)-\frac{n}{2}log(2\pi)\\
&-\frac{\alpha}{2}m^{T}m-\frac{\beta}{2}{||Fm-y||}^{2}-\frac{1}{2}log(|A|)\\
\end{split}
\end{equation}
Following, the maximized value in Equation \ref{eq:LogME} with optimized variances $\alpha^{*}$ and $\beta^{*}$ is the \textbf{LogME} score.

\section{Methodology}\label{sec:method}
\subsection{Log-Likelihood Ratio}\label{sec:llr}
To evaluate the comparative performance of the linear probing (LP) model $\theta$ and the visual prompting (VP) model $\theta_{p}$ with the given dataset, we utilize the log-likelihood ratio (LLR) method described in Equation \ref{eq:LLR}. The terms $p_{\theta}$ and $p_{\theta_{p}}$ denote the maximum likelihoods of LP and VP, respectively. By decomposing the input $x$ into components $x_{ID}$ and $x_{OOD}$, we can analyze the distinct impacts of visual prompts on ID and OOD inputs. As discussed in Section \ref{sec:intro}, in ID datasets, prompts may disrupt the dominant features from the clean images, resulting in an LLR score below 0 (\ref{eq:LLR_ID}). Conversely, for OOD datasets, prompts enhance the model's recognition ability by providing crucial features, yielding an LLR score above 0 (\ref{eq:LLR_OOD}).
\begin{equation} \label{eq:LLR}
\begin{split}
&LLR(x) :=\log \frac{p_{\theta _{p}}(x)}{p_{\theta}(x)}=\log\text{ } p_{\theta _{p}}(x)-log\text{ }p_{\theta}(x)\\
&:=\log(p_{\theta _{p}}(x_{ID}) p_{\theta _{p}}(x_{OOD}))-\log(p_{\theta}(x_{ID}) p_{\theta}(x_{OOD}))
\end{split}
\end{equation}
\begin{equation}\tag{LLR: Dominant ID Features} \label{eq:LLR_ID}
\begin{split}
&LLR(x) \sim \log\text{ } p_{\theta _{p}}(x_{ID})-\log\text{ } p_{\theta}(x_{ID})<0 
\end{split}
\end{equation}
\begin{equation} \tag{LLR: Dominant OOD Features} \label{eq:LLR_OOD}
\begin{split}
&LLR(x) \sim \log\text{ } p_{\theta _{p}}(x_{OOD})-\log\text{ } p_{\theta}(x_{OOD})>0 
\end{split}
\end{equation}

\subsection{LogME Evidence and Visual Prompting Evidence} 
To obtain the maximum log-likelihood $\log p_{\theta}$, we follow the method in LogME \cite{you2021logme}, as described in Section \ref{related_work:logme}. When considering the VP model, the impact of visual prompts is reflected in the feature matrix as $F(\delta)$. Therefore, $\log p_{\theta_{p}}$ can be obtained from equation \textbf{\ref{eq:LogME_VP}
}, which involves computing the expectation value of the evidence with respect to the visual prompt $\delta$.
\begin{equation} \tag{LogME-VP} \label{eq:LogME_VP}
\begin{split}
p(y|F,\alpha,\beta)
&=\int_{w} p(w|\alpha) \VP{\int_{\delta} p(\delta)} p(y|\VP{F(\delta)}, w,\beta, \VP{\delta})\VP{d\delta}dw\\
&=\VP{\int_{\delta} p(\delta)} \underbrace{\int_{w} p(w|\alpha) p(y|\VP{F(\delta)}, w,\beta, \VP{\delta})dw}_\text{Evidence} \VP{d\delta}\\
& =\mathbb{E}_{\delta}[\text{Evidence}(\VP{\delta}) ];\\
\log(p(y|F,\alpha,\beta))
& =\mathbb{E}_{\delta}[\log\big(\text{Evidence}(\VP{\delta})\big)]
\end{split}
\end{equation}

At this stage, once $\log p_{\theta {p}}$ and $\log p_{\theta}$ are calculated, subtracting the two provides the LLR scores in Equation \ref{eq:LLR}.

\subsection{Visual Prompt Approximation}\label{sec:vp_approximation}
\begin{table}[t]
\caption{\textbf{Comparison of Linear Probing (LP), Full Fine-Tuning (FF), and the LLR Score with Simulated Visual Prompts.} Using the EuroSAT dataset and CLIP as the pre-trained model, the table presents execution time, trainable parameter size, and training dataset size for each method.}
\label{tab:computation}
\centering
\begin{tabular}{ccccccc}
\toprule
\begin{tabular}[c]{@{}c@{}}Experimental \\ Info.\end{tabular} &
  LP &
  FF &
  \begin{tabular}[c]{@{}c@{}}LLR (a)\\ Gaussian\end{tabular} &
  \begin{tabular}[c]{@{}c@{}}LLR (b)\\ Gradient\end{tabular} &
  \begin{tabular}[c]{@{}c@{}}LLR (c)\\ Mini-FT\end{tabular} &
  \begin{tabular}[c]{@{}c@{}}LLR (d)\\ Mini-FT-5R\end{tabular} \\ \midrule
\begin{tabular}[c]{@{}c@{}}Execution \\ Time (second)\end{tabular}               & 2370   & 3081   & 22 & 27          & 23          & 33          \\ \midrule
\begin{tabular}[c]{@{}c@{}}Trainable \\ Parameter \\ Size (Million)\end{tabular} & 0.005  & 151.28 & 0  & 0           & 0.15        & 0.15        \\ \midrule
\begin{tabular}[c]{@{}c@{}}Training \\ Dataset Size\end{tabular}                 & 13,500 & 13,500 & 0  & $\sim$1,000 & $\sim$1,000 & $\sim$1,000 \\ \bottomrule
\end{tabular}
\end{table}
Given the vastness of the $\mathbb{R}^{3\times N\times N}$ prompt space for images, exhaustively exploring all visual prompts to compute the expectation value in \ref{eq:LogME_VP} is impractical. Hence, we explore different methods for simulating prompt distributions. These methods can be categorized into two approaches: the training-free approach and the mini-finetuning approach.

(i) Training-Free Approach:
We employ two distributions to model visual prompts. The first one is an isotropic multivariate Gaussian distribution, which simulates prompts $\delta \sim N(0, \gamma I)$ with standard deviation $\gamma$. The second method is the gradient approximation approach, where prompts' gradients are computed from a small subset ($\sim$1k) of samples $\{x_{i}\}^{n}_{i=1}$   \cite{krishnamachari2023fourier} and the loss function $E$ is minimized via Taylor expansion approximation \cite{tang2024neural,hsiung2023nctv} (see Equation \ref{eq:taylor}).
\begin{equation}  \label{eq:taylor}
\begin{split}
&E(x+\delta) \approx E(x) + \delta^{T}\bigtriangledown E(x)\\
&{\nabla E(x)}_{i} = g_{i}\text{, the gradient respected to pixel $i$}.\\
&\text{minimize: }\delta^{T}\nabla E(x)=\delta^{T}\textbf{g} = \sum \delta_{i}g_{i}\\
&\Rightarrow \left \{
\begin{aligned}
&g_{i}\geq0\rightarrow \delta_{i} = 0\\
&g_{i}<0\rightarrow \delta_{i} = 1 \text{, for pixel value $\delta_{i}$} \in [0,1].
\end{aligned}
\right.
\end{split}
\end{equation} 
(ii) Mini-Finetuning:
We select a small subset of samples and exclusively update the prompts. This process is conducted under two configurations: 1\&5-epoch tuning.

\begin{figure}[t]
    \centering
    \includegraphics[width=\linewidth]{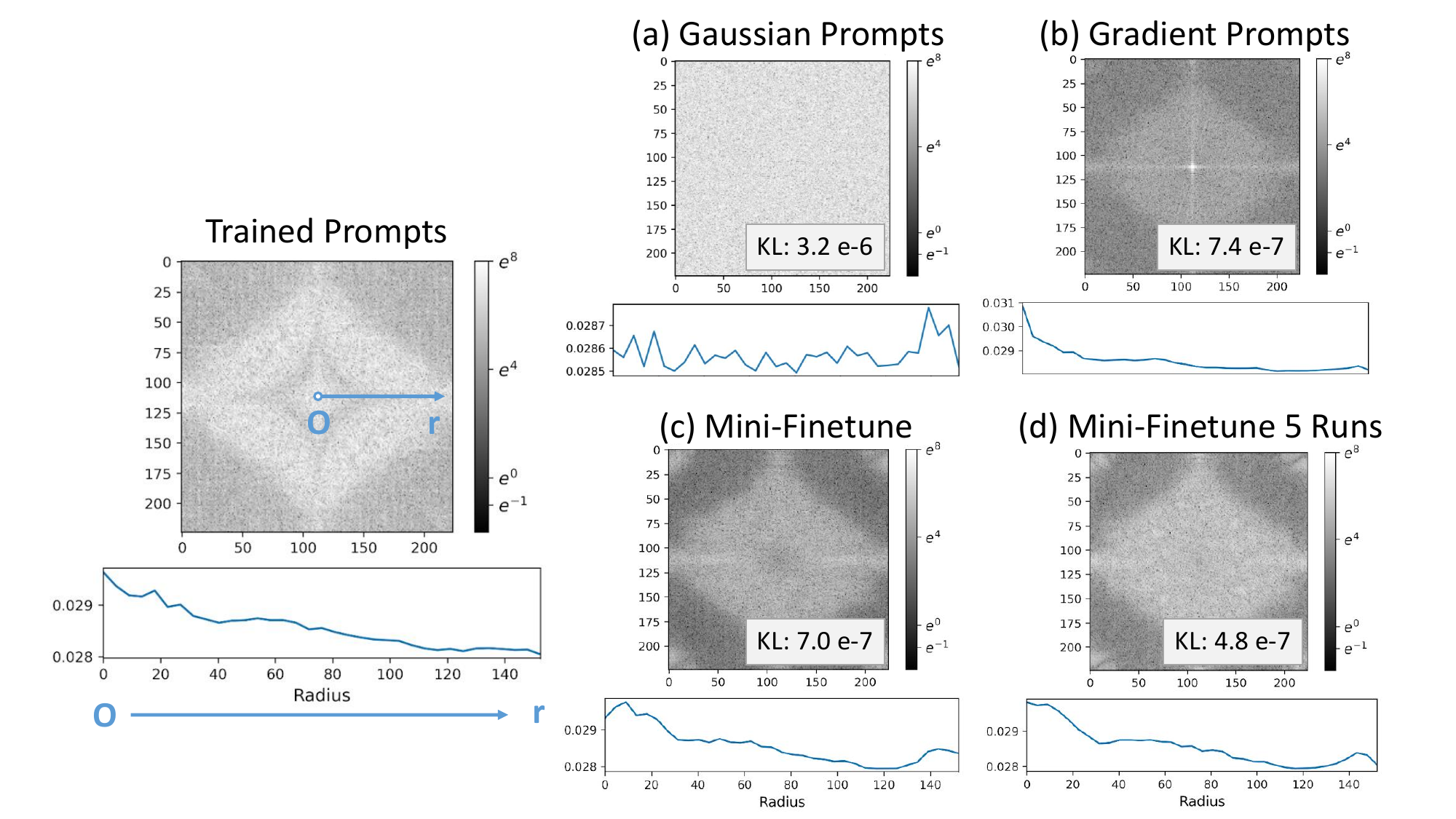}
    \caption {\textbf{The Similarity of Visual Prompts on CLIP (ViT/B-32).} Various types of prompts are presented in the frequency domain, along with line plots of the average values with radii. The similarity between simulated and trained visual prompts is evaluated using KL divergence \cite{kullback1951information}.}
    \label{fig:prompt_similarity}
\end{figure}

Fig. \ref{fig:prompt_similarity} (a) to (d) shows the distribution of the four simulated prompts in the frequency domain. From method (a) to (b) and referring to Table \ref{tab:computation}, there is an increase in data utilization, moving from a data-free approach to a limited-data setting. Subsequently, from (c) to (d), the computation increases fivefold for the tuning process. This demonstrates that with increased resources, whether data or computation, the simulated prompts can approximate the trained ones more closely, validated by the decrease in KL divergence between their distributions.

\section{Experimental Results}\label{sec:exp}
% Please add the following requibrickred packages to your document preamble:
\begin{table*}[t]
\caption{\textbf{LLR Sorting Results}. The evaluated metrics include \color[HTML]{6AA84F}{Kendall's $\tau$ in green}\color{black}, \color[HTML]{CE2029}{Spearman's $\rho$ in red}\color{black}, and \color[HTML]{26619C}{LLR-Acc (\%) in blue}\color{black}. The highest score is \textbf{\underline{underlined}} (excluding scores from trained prompts). The baseline methods (see Appendix \ref{apx:baseline}) include the confidence score \cite{hendrycks2016baseline}, the ODIN confidence score \cite{liang2017enhancing}, and AUROC with Mahalanobis Distance \cite{lee2018simple}.}
\label{tab:all_score}
\centering
\resizebox{\textwidth}{!}{
\begin{tabular}{c|cccccc|ccc}
\toprule
 &
  \begin{tabular}[c]{@{}c@{}}Trained\\ Prompts\end{tabular} &
  \begin{tabular}[c]{@{}c@{}}Without\\ Prompts\end{tabular} &
  Gaussian &
  Gradient &
  Mini-Finetune &
  \begin{tabular}[c]{@{}c@{}}Mini-Finetune\\ 5 Rounds\end{tabular} &
  \begin{tabular}[c]{@{}c@{}}Confidence\\ Score\end{tabular} &
  \begin{tabular}[c]{@{}c@{}}ODIN\\ Confidence\end{tabular} &
  \begin{tabular}[c]{@{}c@{}}Mahalanobis Distance\\ AUROC\end{tabular} \\ \midrule
ResNet18 &
  {\begin{tabular}[c]{@{}c@{}} \color[HTML]{6AA84F}0.61\\ \color[HTML]{CE2029}0.77\\ \color[HTML]{26619C}75.0\end{tabular}} &
  {\begin{tabular}[c]{@{}c@{}} \color[HTML]{6AA84F}\textbf{\underline{0.67}}\\ \color[HTML]{CE2029} \textbf{\underline{0.83}}\\ \color[HTML]{26619C}41.7\end{tabular}} &
  {\begin{tabular}[c]{@{}c@{}} \color[HTML]{6AA84F}0.48\\ \color[HTML]{CE2029}0.65\\ \color[HTML]{26619C}\textbf{\underline{58.3}}\end{tabular}} &
  {\begin{tabular}[c]{@{}c@{}} \color[HTML]{6AA84F}0.52\\ \color[HTML]{CE2029}0.71\\ \color[HTML]{26619C}\textbf{\underline{58.3}}\end{tabular}} &
  {\begin{tabular}[c]{@{}c@{}} \color[HTML]{6AA84F}0.42\\ \color[HTML]{CE2029}0.62\\ \color[HTML]{26619C}\textbf{\underline{58.3}}\end{tabular}} &
  {\begin{tabular}[c]{@{}c@{}} \color[HTML]{6AA84F}0.45\\ \color[HTML]{CE2029}0.61\\ \color[HTML]{26619C}\textbf{\underline{58.3}}\end{tabular}} &
  {\begin{tabular}[c]{@{}c@{}} \color[HTML]{6AA84F}0.27\\ \color[HTML]{CE2029}0.41\end{tabular}} &
  {\begin{tabular}[c]{@{}c@{}} \color[HTML]{6AA84F}0.21\\ \color[HTML]{CE2029}0.31\end{tabular}} &
  {\begin{tabular}[c]{@{}c@{}} \color[HTML]{6AA84F}0.52\\ \color[HTML]{CE2029}0.69\end{tabular}} \\ \midrule
ResNext-IG &
  {\begin{tabular}[c]{@{}c@{}} \color[HTML]{6AA84F}0.58\\ \color[HTML]{CE2029}0.78\\ \color[HTML]{26619C}83.3\end{tabular}} &
  {\begin{tabular}[c]{@{}c@{}} \color[HTML]{6AA84F}0.18\\ \color[HTML]{CE2029}0.24\\ \color[HTML]{26619C}50.0\end{tabular}} &
  {\begin{tabular}[c]{@{}c@{}} \color[HTML]{6AA84F}0.61\\ \color[HTML]{CE2029}0.73\\ \color[HTML]{26619C}66.7\end{tabular}} &
  {\begin{tabular}[c]{@{}c@{}} \color[HTML]{6AA84F}\textbf{\underline{0.64}}\\ \color[HTML]{CE2029}\textbf{\underline{0.80}}\\ \color[HTML]{26619C}\textbf{\underline{83.3}}\end{tabular}} &
  {\begin{tabular}[c]{@{}c@{}} \color[HTML]{6AA84F}0.33\\ \color[HTML]{CE2029}0.48\\ \color[HTML]{26619C}58.3\end{tabular}} &
  {\begin{tabular}[c]{@{}c@{}} \color[HTML]{6AA84F}0.27\\ \color[HTML]{CE2029}0.38\\ \color[HTML]{26619C}58.3\end{tabular}} &
  {\begin{tabular}[c]{@{}c@{}} \color[HTML]{6AA84F}0.52\\ \color[HTML]{CE2029}0.70\end{tabular}} &
  {\begin{tabular}[c]{@{}c@{}} \color[HTML]{6AA84F}0.33\\ \color[HTML]{CE2029}0.47\end{tabular}} &
  {\begin{tabular}[c]{@{}c@{}} \color[HTML]{6AA84F}0.38\\  \color[HTML]{CE2029}0.60\end{tabular}} \\ \midrule
ViT-B-16 &
  {\begin{tabular}[c]{@{}c@{}} \color[HTML]{6AA84F}0.61\\ \color[HTML]{CE2029}0.81\\ \color[HTML]{26619C}66.7\end{tabular}} &
  {\begin{tabular}[c]{@{}c@{}} \color[HTML]{6AA84F}0.42\\ \color[HTML]{CE2029}0.66\\ \color[HTML]{26619C}75.0\end{tabular}} &
  {\begin{tabular}[c]{@{}c@{}} \color[HTML]{6AA84F}0.58\\ \color[HTML]{CE2029}0.77\\ \color[HTML]{26619C}\textbf{\underline{91.7}}\end{tabular}} &
  {\begin{tabular}[c]{@{}c@{}} \color[HTML]{6AA84F}\textbf{\underline{0.64}}\\ \color[HTML]{CE2029}\textbf{\underline{0.82}}\\ \color[HTML]{26619C}\textbf{\underline{91.7}}\end{tabular}} &
  {\begin{tabular}[c]{@{}c@{}} \color[HTML]{6AA84F}0.48\\ \color[HTML]{CE2029}0.63\\ \color[HTML]{26619C}\textbf{\underline{91.7}}\end{tabular}} &
  {\begin{tabular}[c]{@{}c@{}} \color[HTML]{6AA84F}0.48\\ \color[HTML]{CE2029}0.64\\ \color[HTML]{26619C}\textbf{\underline{91.7}}\end{tabular}} &
  {\begin{tabular}[c]{@{}c@{}} \color[HTML]{6AA84F}0.18\\ \color[HTML]{CE2029}0.18\end{tabular}} &
  {\begin{tabular}[c]{@{}c@{}} \color[HTML]{6AA84F}-0.18\\ \color[HTML]{CE2029}-0.31\end{tabular}} &
  {\begin{tabular}[c]{@{}c@{}} \color[HTML]{6AA84F}0.61\\ \color[HTML]{CE2029}0.77\end{tabular}} \\ \midrule
Swin-T &
  {\begin{tabular}[c]{@{}c@{}} \color[HTML]{6AA84F}0.58\\ \color[HTML]{CE2029}0.80\\ \color[HTML]{26619C}91.7\end{tabular}} &
  {\begin{tabular}[c]{@{}c@{}} \color[HTML]{6AA84F}0.39\\ \color[HTML]{CE2029}0.56\\ \color[HTML]{26619C}50.0\end{tabular}} &
  {\begin{tabular}[c]{@{}c@{}} \color[HTML]{6AA84F}0.42\\ \color[HTML]{CE2029}0.67\\ \color[HTML]{26619C}83.3\end{tabular}} &
  {\begin{tabular}[c]{@{}c@{}} \color[HTML]{6AA84F}0.42\\ \color[HTML]{CE2029}0.63\\ \color[HTML]{26619C}83.3\end{tabular}} &
  {\begin{tabular}[c]{@{}c@{}} \color[HTML]{6AA84F}0.48\\ \color[HTML]{CE2029}0.69\\ \color[HTML]{26619C}83.3\end{tabular}} &
  {\begin{tabular}[c]{@{}c@{}} \color[HTML]{6AA84F}\textbf{\underline{0.70}}\\ \color[HTML]{CE2029}\textbf{\underline{0.81}}\\ \color[HTML]{26619C}\textbf{\underline{91.7}}\end{tabular}} &
  {\begin{tabular}[c]{@{}c@{}} \color[HTML]{6AA84F}0.48\\ \color[HTML]{CE2029}0.73\end{tabular}} &
  {\begin{tabular}[c]{@{}c@{}} \color[HTML]{6AA84F}0.39\\ \color[HTML]{CE2029}0.59\end{tabular}} &
  {\begin{tabular}[c]{@{}c@{}} \color[HTML]{6AA84F}0.52\\ \color[HTML]{CE2029}0.73\end{tabular}} \\ \midrule
CLIP(VIT/B-32) &
  {\begin{tabular}[c]{@{}c@{}} \color[HTML]{6AA84F}0.64\\ \color[HTML]{CE2029}0.84\\ \color[HTML]{26619C}83.3\end{tabular}} &
  {\begin{tabular}[c]{@{}c@{}} \color[HTML]{6AA84F}0.27\\ \color[HTML]{CE2029}0.38\\ \color[HTML]{26619C}50.0\end{tabular}} &
  {\begin{tabular}[c]{@{}c@{}} \color[HTML]{6AA84F}0.55\\ \color[HTML]{CE2029}0.73\\ \color[HTML]{26619C}66.7\end{tabular}} &
  {\begin{tabular}[c]{@{}c@{}} \color[HTML]{6AA84F}0.55\\ \color[HTML]{CE2029}0.69\\ \color[HTML]{26619C}66.7\end{tabular}} &
  {\begin{tabular}[c]{@{}c@{}} \color[HTML]{6AA84F}\textbf{\underline{0.61}}\\ \color[HTML]{CE2029}\textbf{\underline{0.79}}\\ \color[HTML]{26619C}66.7\end{tabular}} &
  {\begin{tabular}[c]{@{}c@{}} \color[HTML]{6AA84F}0.52\\ \color[HTML]{CE2029}0.71\\ \color[HTML]{26619C}\textbf{\underline{75.0}}\end{tabular}} &
  {\begin{tabular}[c]{@{}c@{}} \color[HTML]{6AA84F}0.39\\ \color[HTML]{CE2029}0.62\end{tabular}} &
  {\begin{tabular}[c]{@{}c@{}} \color[HTML]{6AA84F}0.15\\ \color[HTML]{CE2029}0.15\end{tabular}} &
  {\begin{tabular}[c]{@{}c@{}} \color[HTML]{6AA84F}0.45\\ \color[HTML]{CE2029}0.61\end{tabular}} \\ \bottomrule
\end{tabular}
}
\vspace{-3mm}
\end{table*}
In this section, we will demonstrate the efficacy of the proposed LLR scores and simulated prompts. These techniques will be utilized to rank datasets according to the accuracy gains achieved by VP compared to LP.

\begin{figure}[ht]
    \centering
    \includegraphics[width=0.7\linewidth]{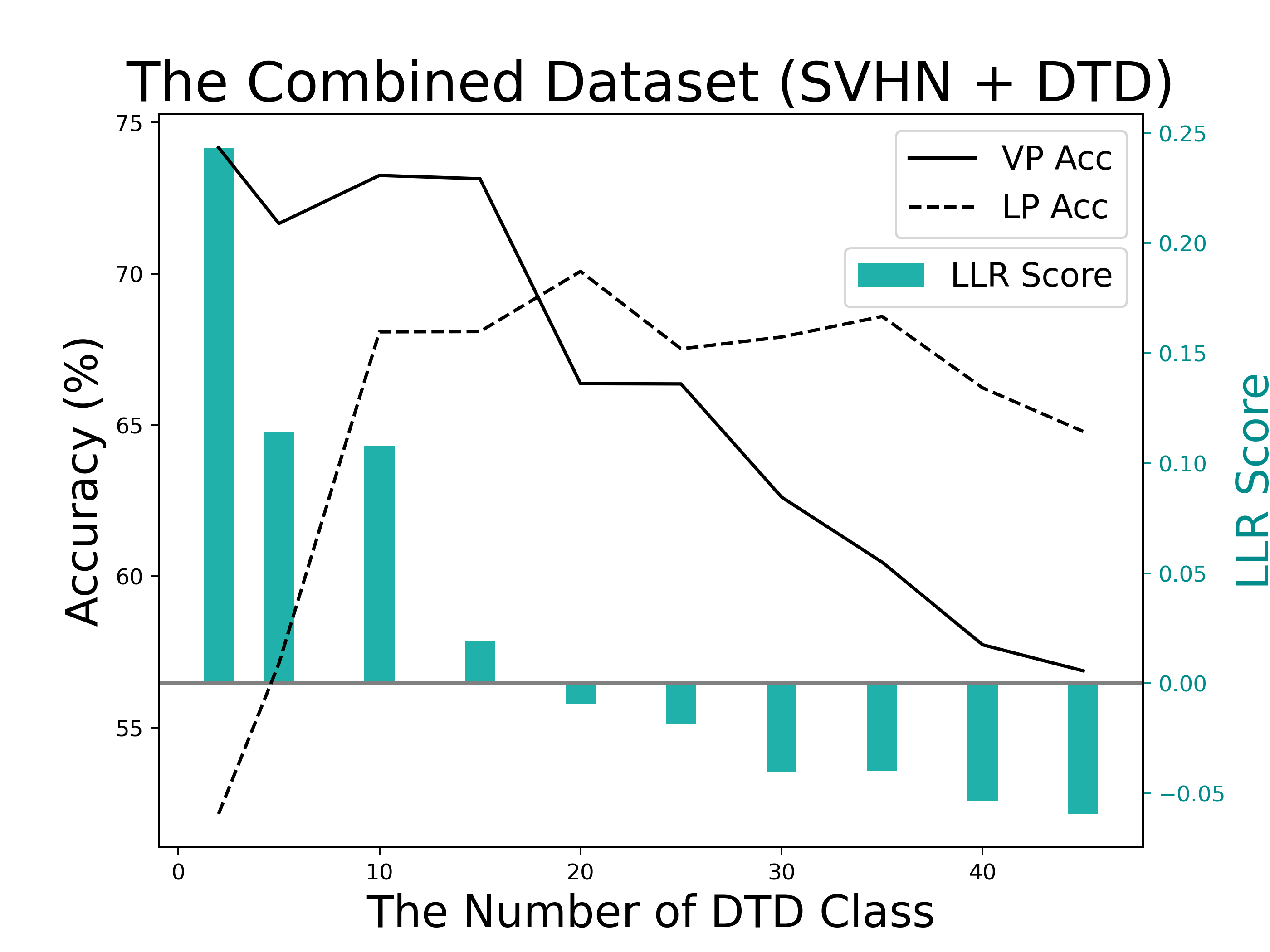}
    \caption {\textbf{Accuracy and LLR on the Combined Datasets Using CLIP (ViT-B/32).} SVHN is considered more OOD, while DTD tends to be ID. There are 10 classes in SVHN, and we gradually increase the number of classes in DTD from 2 to 45 to obtain different ID/OOD proportions. }
    \label{fig:combined_dataset}
\end{figure}

\subsection{The Effectiveness of LLR and Simulated Prompts}
LLR scores in Section \ref{sec:llr} are utilized to differentiate the impact of visual prompts on ID/OOD datasets. In Fig. \ref{fig:combined_dataset}, using a mixed dataset as an example, we observe a close correlation between LLR scores and actual accuracy gains, indicating the method effectively identifies whether VP or LP provides an advantage. A positive LLR score suggests that datasets (leaning towards OOD) benefit more from VP training, while a negative LLR score indicates that datasets (leaning towards ID) are more suitable for LP training.

\begin{figure}[h]
    \centering
    \includegraphics[width=0.9\linewidth]{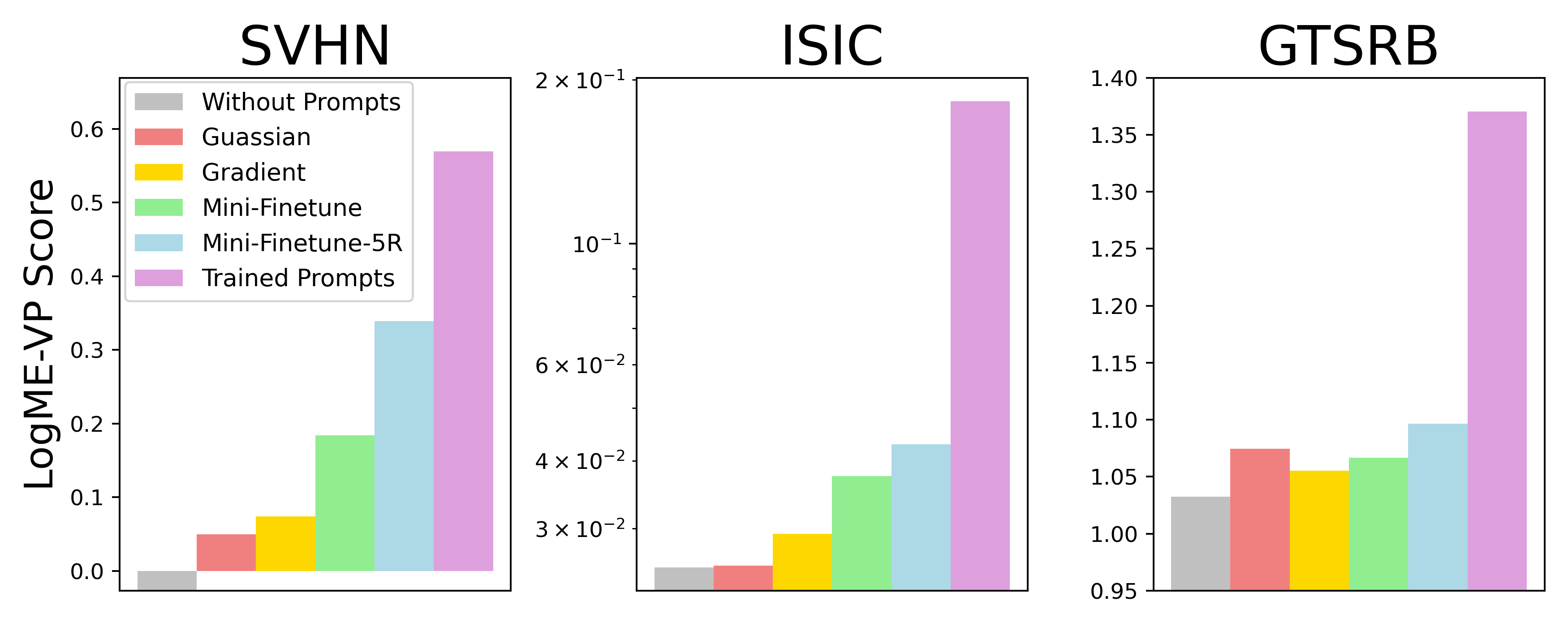}
    \caption {\textbf{The \textbf{\ref{eq:LogME_VP}} Scores with Prompts.} The plot shows the \ref{eq:LogME_VP} scores obtained from CLIP (ViT-B/32)  with various input prompts, including without prompts, Gaussian prompts, gradient prompts, mini-finetuning prompts, and well-trained prompts.}
    \label{fig:score_w_prompt}
\end{figure}

Furthermore, we have validated the efficacy of the simulation methods proposed in Section \ref{sec:vp_approximation}. From two aspects -- the distributional fidelity indicated by decreasing KL divergence (Fig. \ref{fig:prompt_similarity}) and increasing \ref{eq:LogME_VP} scores (Fig. \ref{fig:score_w_prompt})—show that the simulated prompts progressively converge towards those obtained from the trained prompts.

\subsection{The Sorting Results with Diverse Datasets} \label{sec:llr_sorting}
We applied the proposed LLR method to 12 downstream datasets and 5 pre-trained models (see Appendix \ref{apx:datasetmodel}). To evaluate the performance of the LLR scores, we used ranking coefficients such as Kendall’s $\tau$ \cite{kendall1938new} and Spearman’s $\rho$ \cite{spearman1904proof} to assess the alignment between LLR scores and accuracy gains. Additionally, $\text{LLR-Acc}=\frac{\textbf{TP}+\textbf{FN}}{n}$ measures the proportion of datasets that remain in the correct quadrant, with \textbf{TP} representing true positives (positive gains with positive LLR scores) and \textbf{FN} representing false negatives (negative gains with negative LLR scores). The performance gains are obtained from \cite{bahng2022exploring} for LP and from AutoVP \cite{tsao2024autovp} for VP.

\begin{figure}[h]
    \centering
    \includegraphics[width=0.95\linewidth]{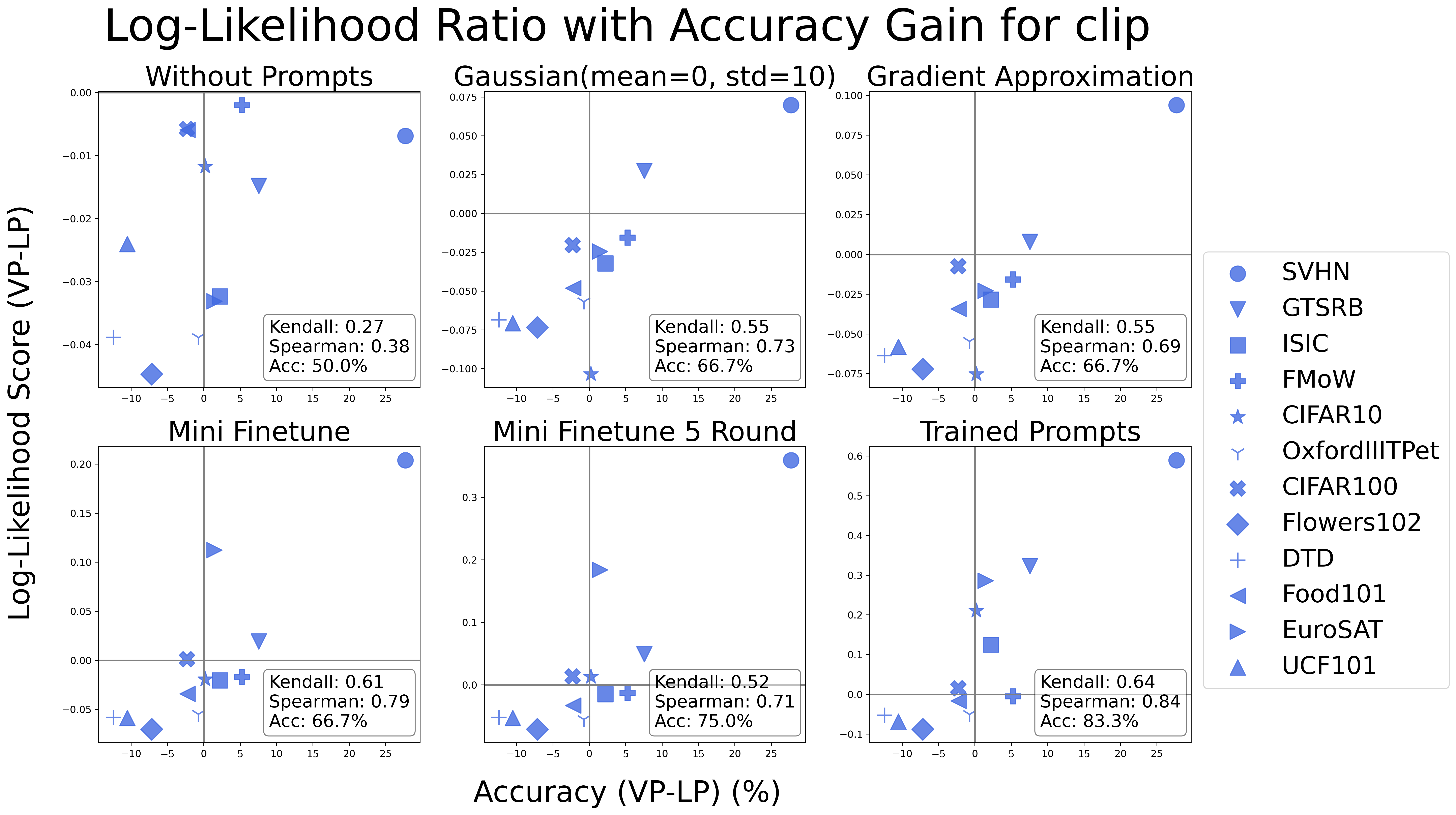}
    \caption {\textbf{The Sorting Results with Simulated Prompts.} The accuracy gains obtained from CLIP (ViT-B/32) are sorted by the LLR scores.}
    \label{fig:clip_logme}
\end{figure}

The visualization of the sorting results with CLIP is presented in Fig. \ref{fig:clip_logme}. It is clear that the inclusion of simulated prompts significantly enhances both the ranking scores and LLR-Acc compared to the scenario without prompts. Additionally, the overall performance are detailed in Table \ref{tab:all_score}. Our LLR method outperforms all baseline methods for OOD detection, offering a more accurate estimation of accuracy gains. This enhanced performance is attributed to precise likelihood calculations and the incorporation of prompt approximations that closely align with the design of visual prompting.
Based on these results, utilizing LLR scores facilitates a more accurate determination of the appropriate training approach. Specifically, datasets with higher LLR scores are more suitable for visual prompting, while lower LLR scores indicate a better fit for linear probing.

\section{Conclusion}\label{sec:conclusions}
This paper proposes an LLR score using visual prompt approximation methods to evaluate the advantage of VP over LP. The LLR scores demonstrate improved ranking with actual accuracy gains and reliable performance prediction, effectively correlating with the proportion of OOD data in the dataset. Consequently, this approach serves as a valuable precursor for training in transfer learning, significantly reducing the time for exploring different fine-tuning methods.

\bibliographystyle{plainnat}
\bibliography{example_paper}

\newpage
\appendix

\section*{Appendix}\label{sec:appendix}
\section{Datasets and Pre-Trained Models}\label{apx:datasetmodel}
We employed 12 datasets for LLR sorting in Section \ref{sec:llr_sorting}. Detailed information is provided in Table \ref{tab:data_info}. We utilized five pre-trained models: the convolutional-based models (ResNet18 \cite{he2016deep} and ResNext-IG \cite{mahajan2018exploring}), vision transformers (ViT-B-16 \cite{dosovitskiy2020image} and Swin-T \cite{liu2021swin}), and the multi-modal model (CLIP \cite{radford2021learning}). ResNext-IG is pre-trained on billions of Instagram images with 1000 classes. CLIP is trained using image-text pairs as input, enabling zero-shot prediction without a fixed predicted class number. The remaining models are pre-trained on ImageNet-1K \cite{imagenet15russakovsky}, which includes 1000 distinct labels.
\begin{table}[h]
\centering
\caption{\textbf{Dataset Information.} The table shows the number of classes for each dataset and the respective LP \cite{bahng2022exploring} and VP \cite{tsao2024autovp} accuracy (\%) with CLIP.}
\label{tab:data_info}
\renewcommand{\arraystretch}{1.25}
\begin{tabular}{c|c|c|c}
\hline
Dataset    & Class Number & LP Accuracy & VP Accuracy\\ \hline
SVHN \cite{netzer2011reading}      & 10   &  65.4  & 93.1 \\ \hline
EuroSAT \cite{helber2019eurosat}   & 10 & 95.3 & 96.8 \\ \hline
Flowers102 \cite{Nilsback08} & 102    & 96.9 &  89.7 \\ \hline
CIFAR100 \cite{Krizhevsky09learningmultiple}  & 100 & 80.0 &  77.7   \\ \hline
UCF101 \cite{soomro2012ucf101}     & 101  & 83.3 & 72.8    \\ \hline
DTD \cite{cimpoi14describing}       & 47 & 74.6 &  62.2   \\ \hline
FMoW \cite{christie2018functional}      & 62  & 36.3 & 41.5   \\ \hline
GTSRB \cite{Houben-IJCNN-2013}     & 43  & 85.8 &  93.4    \\ \hline
CIFAR10 \cite{Krizhevsky09learningmultiple}    & 10 & 95.0 &   95.2   \\ \hline
Food101 \cite{bossard14}   & 101      &  84.6 &  82.4  \\ \hline
OxfordIIITPet \cite{parkhi2012cats} & 37 & 89.2 & 88.4  \\ \hline
ISIC \cite{codella2019skin,tschandl2018ham10000}  & 7 & 71.9 & 74.1   \\
\hline
\end{tabular}
\end{table}

\section{VP and LP Performance}
The performance gains are obtained from the accuracy difference between VP and LP. Therefore, we use the LP accuracy from \cite{bahng2022exploring}, and VP accuracy from AutoVP \cite{tsao2024autovp}. For pre-trained models not included in these works, we trained them ourselves using the same settings. In VP, the output mapping is FullyMap (as described in AutoVP Section 3), with prompt frame sizes of 48 for SVHN and 16 for the other datasets.

\section{Feature Extraction for Evidence Score Calculation}
\begin{figure}[h]
    \centering
    \includegraphics[width=\linewidth]{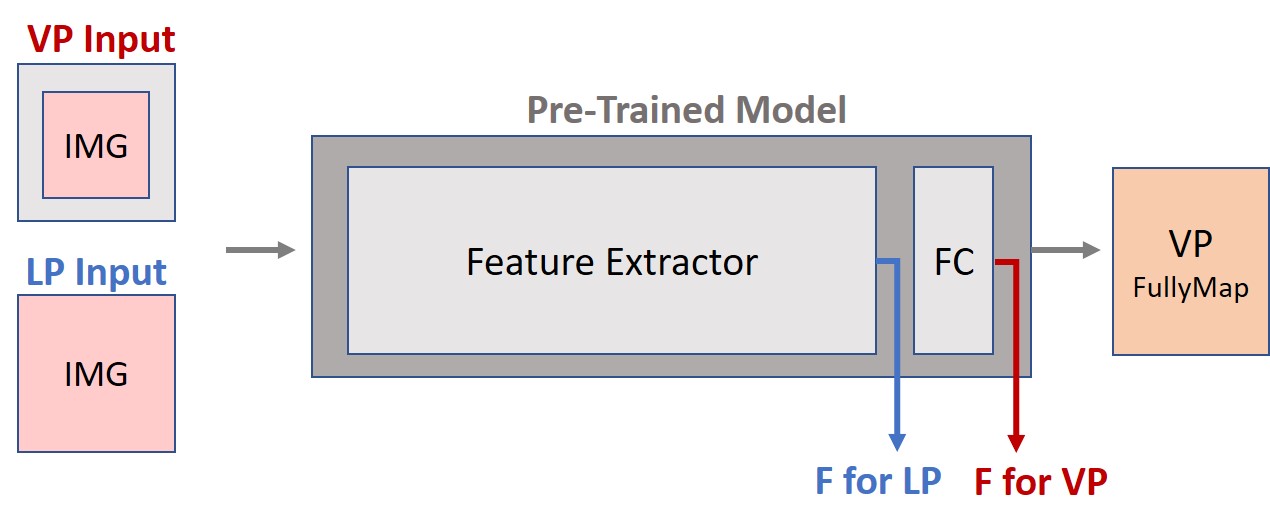}
    \caption {\textbf{Feature Extraction for VP and LP.} The symbols marked in blue and red represent the features used for computing the evidence score for LP and VP, respectively, and are extracted from different layers of the pre-trained model.}
    \label{fig:feature_lp_vp}
\end{figure}

When calculating the LogME scores, the feature $F$ is derived from the output of the pre-trained model, which serves as a feature extractor. Due to the differences in the frameworks of VP and LP, the features extracted for computing the evidence scores vary. Details can be found in Fig. \ref{fig:feature_lp_vp}. For LP, the pre-trained model's linear classifier is adjusted for adaptation, so the feature $F$ is extracted from the output of the pre-trained model. However, since the VP model retains the pre-trained output classifier and obtains a mapping (e.g., FullyMap), the $F(\delta)$ is extracted from the model's linear classifier (i.e. \textbf{FC} in Fig. \ref{fig:feature_lp_vp}). Additionally, the dimensions of $F$ vary depending on the extraction point. Table \ref{tab:feature_dim} show the dimensions of $F$ under different frameworks.

\begin{table}[h]
\caption{\textbf{Feature Dimension of VP and LP.} $N$ represents the input sample size. In CLIP, 81 is the number of text templates used, and $Cls\_Num$ denotes the number of classes in the dataset.}\label{tab:feature_dim}
\centering
\renewcommand{\arraystretch}{1.25}
\begin{tabular}{c|c|c|c|c|c}
\hline
Framework            & ResNet18 & IG       & ViT-B    & Swin-T   & CLIP-VIT/B-32   \\ \hline
\begin{tabular}[c]{@{}c@{}}VP Feature\\ Dimension\end{tabular} & (N,1000) & (N,1000) & (N,1000) & (N,1000) & (N,81*$Cls\_Num$) \\ \hline
\begin{tabular}[c]{@{}c@{}}LP Feature\\ Dimension\end{tabular}  & (N,512)  & (N,2048) & (N,768)  & (N,768)  & (N,512)         \\ \hline
\end{tabular}
\end{table}

\section{Sorting Metrics} \label{apx:sortingmetric}

In Section\ref{sec:llr_sorting}, we introduce three metrics to evaluate the performance of the LLR sorting results. The ranking metrics, \ref{eq:kendall} and \ref{eq:Spearman}, range from -1 to 1, where 1 indicates a perfectly correct ranking, and -1 indicates a completely reversed ranking. Kendall's $\tau$ serves as a more rigorous metric, necessitating verification of the correct ordering for each pair of items, whereas Spearman's correlation is comparatively less stringent, derived from the summation of the squared differences of the ranks.

% Kendall
\begin{equation}
    \tau(\textbf{X},\textbf{Y})=\frac{2}{n(n-1)}\sum_{i<j}sgn(x_{i}-x_{j})sgn(y_{i}-y_{j}) \tag{Kendall's $\tau$ coefficient} \label{eq:kendall}
\end{equation}

% Spearman
\begin{equation}
    \rho(\textbf{X},\textbf{Y})=1-\frac{6\sum_{i}d_{i}^{2}}{n(n^{2}-1)}\text{, where }d_{i}=Rank(x_{i})-Rank(y_{i}) \tag{Spearman's $\rho$ coefficient} \label{eq:Spearman}
\end{equation}

The final metric, as shown in Equation \ref{eq:LLR-ACC}, calculates the ratio of true positives (TP) and false negatives (FN) to the total data point $n$. When the LLR score is accurately estimated, positive LLR correspond to positive accuracy gains (TP points), while negative LLR correspond to negative accuracy gains (FN points).
\begin{equation}
    Acc(\textbf{X},\textbf{Y})=\frac{\textbf{TP}+\textbf{FN}}{n} \tag{LLR-Accuracy} \label{eq:LLR-ACC}
\end{equation}

\section{OOD Detection Baselines} \label{apx:baseline}
In Section \ref{sec:intro}, we assert that data tending to be OOD may exhibit larger accuracy gains, while data tending to be ID may show smaller or even negative accuracy gains. Therefore, in Section \ref{sec:llr_sorting}, we utilize several OOD detection techniques as baseline methods for comparison with our LLR scores. The OOD detection baselines as detailed below:
\begin{itemize}

\item Confidence score \cite{hendrycks2016baseline}, defined as the maximum class probability from a softmax classifier, is represented as $p(\hat{y}|x) = max_{c}\text{ }p(y=c|x)$, where label $c$ is from $1\sim N$ and $x$ is the input samples. Higher scores indicate ID data, while lower scores indicate OOD data.

\item ODIN confidence \cite{liang2017enhancing} employs techniques such as temperature scaling and adding small perturbations to the input better to separate confidence scores of ID and OOD samples.

\item Mahalanobis distance \cite{lee2018simple}, $M_{\ell} = min_{c} (f_{\ell}(x)-\mu_{\ell,c})^{T} \Sigma_{\ell}^{-1} (f_{\ell}(x)-\mu_{\ell,c})$, computes the closest Gaussian distance from the training classes $c$ by using the features of layer $\ell$ in pre-trained models. We use the output AUROC scores for sorting; higher AUROC indicates more OOD.
\end{itemize}

For methods using confidence scores and ODIN confidence, we add a negative sign before sorting, as higher scores indicate ID data, which is the opposite trend to our LLR perspective.

\section{LogME For Visual Prompting Model Ranking}

\begin{figure}[htbp]
    \centering
    \includegraphics[width=.6\linewidth]{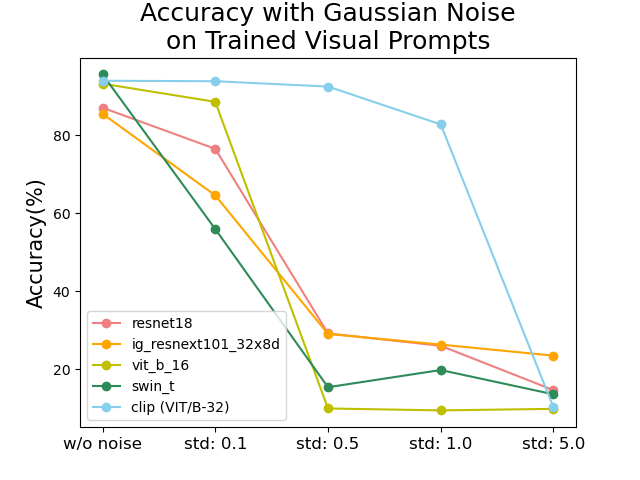}
    \caption {\textbf{Accuracy Decline with Gaussian Noise.} The figure illustrates the accuracy of SVHN across different models with varying levels of Gaussian noise (with a standard deviation ranging from 0.1 to 5.0) are added to the trained visual prompts.}
    \label{fig:svhn_perturb}
\end{figure}

\begin{figure}[t]
    \centering
    \includegraphics[width=\linewidth]{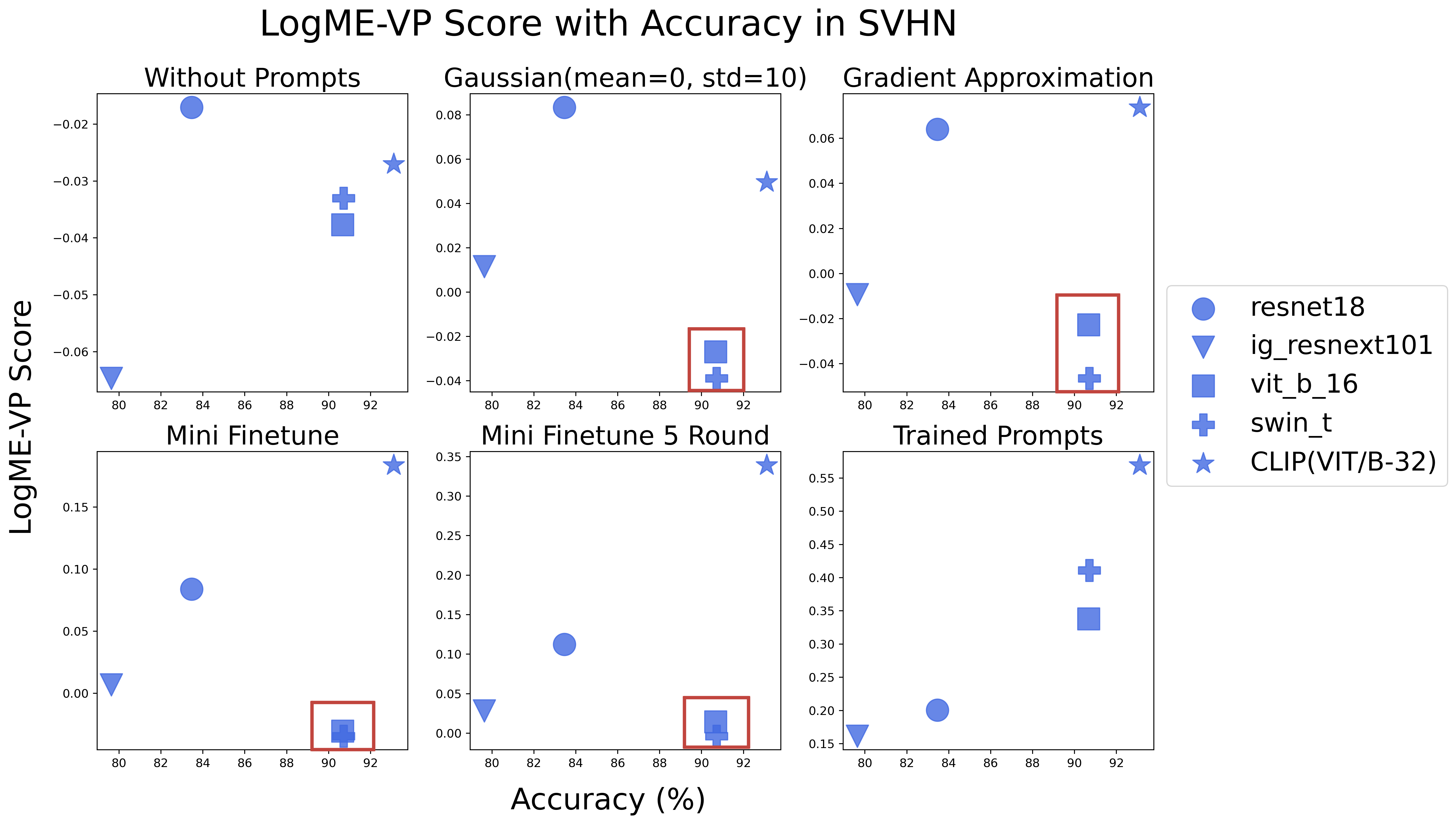}
    \caption {\textbf{LogME-VP Score and Model Performance for SVHN.} The LogME-VP score is computed across several visual prompt settings. The points enclosed in red frames highlight the pre-trained models that fall behind the overall trend.}
    \label{fig:svhn_logme}
\end{figure}

In LogME \cite{you2021logme}, the evidence score is used to rank the transfer performance of pre-trained models. However, this ranking may fail for the visual prompting framework because different models exhibit varying sensitivities to prompts. Fig. \ref{fig:svhn_perturb} illustrates the accuracy decline of different pre-trained models under various levels of Gaussian noise. As shown, when the noise standard deviation increases to 0.5, the accuracy of ViT-B-16 and Swin-t decreases sharply. This indicates that these models are highly sensitive to prompts and require accurate prompts for correct predictions. This result is also reflected in Fig.\ref{fig:svhn_logme}, where ViT-B-16 and Swin-T fail to improve their evidence scores effectively with different simulated prompts. Consequently, their scores fall behind those of other models, making model ranking less effective. Therefore, within the visual prompting framework, we do not rank pre-trained models based on the given dataset. Instead, we compare VP and LP with the same pre-trained model for consistency. (see Section \ref{sec:llr}).

\end{document}